\documentclass{article}

\PassOptionsToPackage{numbers, compress}{natbib}

\usepackage[final]{neurips_2024}

\usepackage{amsmath} 
\usepackage[utf8]{inputenc} 
\usepackage[T1]{fontenc}    
\usepackage{url}            
\usepackage{booktabs}       
\usepackage{amsfonts}       
\usepackage{tablefootnote}
\usepackage{nicefrac}       
\usepackage{microtype}      
\usepackage[numbers]{natbib}
\usepackage[dvipsnames,table,xcdraw]{xcolor}
\definecolor{cvprblue}{rgb}{0.21,0.49,0.74}
\usepackage{graphicx}
\usepackage{caption}
\usepackage{arydshln}
\usepackage{tikz}
\usepackage{wrapfig}
\usepackage{float}
\usepackage{multirow}
\usepackage{pifont}
\usepackage{makecell}
\usepackage{subfigure}
\usepackage[pagebackref,breaklinks,colorlinks,citecolor=cvprblue]{hyperref}

\usepackage{mdframed}
\definecolor{mygray}{gray}{0.97}
\colorlet{shadecolor}{mygray}
\newmdenv[%
  backgroundcolor=mygray, 
  linewidth=0pt
]{newshaded}

\title{Skywork-VL Reward: An Effective  Reward Model for Multimodal Understanding and Reasoning}

\author{
\textnormal{Xiaokun Wang}\thanks{Equal contribution}, \quad \textnormal{Peiyu Wang}\footnotemark[1], \quad   \textnormal{Jiangbo Pei}\footnotemark[1], \quad   \textnormal{Wei Shen}\footnotemark[1], \quad \textnormal{Yi Peng}, \quad \textnormal{Yunzhuo Hao},\\ \quad \textnormal{Weijie Qiu},
 \quad \textnormal{Ai Jian}, \quad \textnormal{Tianyidan Xie}, \quad \textnormal{Xuchen Song}\thanks{Corresponding author}, \quad \textnormal{Yang Liu}, \quad \textnormal{Yahui Zhou}
    \\
    \\
    \textnormal{Skywork AI,  Kunlun Inc.}\\
 {\textnormal{xuchen.song@kunlun-inc.com}}\\
}

\begin{document}

\maketitle

\begin{abstract}
We propose Skywork-VL Reward, a multimodal reward model that provides reward signals for both multimodal understanding and reasoning tasks. Our technical approach comprises two key components: First, we construct a large-scale multimodal preference dataset that covers a wide range of tasks and scenarios, with responses collected from both standard vision-language models (VLMs) and advanced VLM reasoners.  Second, we design a reward model architecture based on Qwen2.5-VL-7B-Instruct, integrating a reward head and applying multi-stage fine-tuning using pairwise ranking loss on pairwise preference data.
Experimental evaluations show that Skywork-VL Reward achieves state-of-the-art results on multimodal VL-RewardBench and exhibits competitive performance on the text-only RewardBench benchmark. Furthermore, preference data constructed based on our Skywork-VL Reward proves highly effective for training Mixed Preference Optimization (MPO), leading to significant improvements in multimodal reasoning capabilities. Our results underscore Skywork-VL Reward as a significant advancement toward general-purpose, reliable reward models for multimodal alignment. Our model has been publicly released to promote transparency and reproducibility\footnote{\href{https://huggingface.co/Skywork/Skywork-VL-Reward-7B}{https://huggingface.co/Skywork/Skywork-VL-Reward-7B}}.
\end{abstract}

\section{Introduction}
Large language models (LLMs) and vision-language models (VLMs) have recently achieved remarkable progress \cite{openai2023gpt4,team2023gemini,openai2024gpt4o,google2024gemini2,team2025kimi,peng2025skywork,chris2025skyworkr1v2multimodalhybrid}, demonstrating impressive capabilities across a wide range of tasks. Despite these advances, aligning their behavior with human preferences remains a significant challenge~\cite{wang2024helpsteer2,yang2024regularizing,peng2025skywork}. Reward models (RMs) have become indispensable in tackling this issue, serving as key components in both the training and inference stages of LLMs and VLMs~\cite{liu2024skyworkrewardbagtricksreward,hao2025mllmsreasonmultimodalityemma,zang2025internlm}.


While reward models for text-only LLMs have been extensively studied, the development of multimodal RMs remains in its early stages, with two major limitations: \textit{Existing multimodal RMs lack generalizability across diverse tasks and struggle to effectively evaluate advanced VLM reasoners with complex inference.} Hence, there is a pressing need for multimodal RMs capable of assessing outputs from both standard VLMs and advanced VLM-based reasoners across diverse domains and tasks.

In this paper, we introduce Skywork-VL Reward, a  multimodal RM designed to serve as a comprehensive and robust evaluator for VLM outputs. Our approach addresses previous limitations in domain coverage and reasoning capacity by incorporating two critical improvements: (i) creating a carefully curated multimodal preference dataset derived from various sources, and (ii) developing a strong base model and training paradigm to enable effective vision-language understanding and reasoning. Specifically, we compile high-quality preference pairs from both publicly available datasets and internal annotations, spanning tasks from basic image descriptions to intricate reasoning scenarios. The collected preference pair includes the image (when applicable), textual prompt, and candidate responses sourced from standard VLMs \cite{bai2025qwen2, chen2023internvl} and advanced VLM reasoners \cite{peng2025skywork}. Building on this dataset, we construct Skywork-VL Reward based on Qwen2.5-VL-7B-Instruct, with an integrated reward head designed to output scalar scores aligned with human preferences. The model is trained using a two-stage training paradigm that combines both pure-text and multimodal data, which enhances its generalization and performance across a wide range of multimodal scenarios. Experimental evaluations confirm that Skywork-VL Reward achieves state-of-the-art results on  VL-RewardBench \cite{li2024vlrewardbenchchallengingbenchmarkvisionlanguage} while maintaining competitive performance in text-only scenarios. 
Furthermore, preference data constructed based on our Skywork-VL Reward proves highly effective when used for training with Mixed Preference Optimization (MPO) \cite{wang2025enhancingreasoningabilitymultimodal}, leading to significant improvements in multimodal reasoning capabilities. 

Our contributions are summarized as follows. First, we introduce Skywork-VL Reward, a multimodal reward model capable of evaluating outputs from both standard VLMs and advanced VLM reasoners across diverse domains and tasks. Second, our model achieves state-of-the-art results on  VL-RewardBench, while maintaining competitive performance in text-only scenarios. Third, preference data generated using Skywork-VL Reward proves highly effective  in MPO training, demonstrating the practical value of our model.

\section{Related Work}\label{sec:related}

\textbf{Reward Models for Text-only Large Language Models.} Reward modeling has become a cornerstone of aligning LLM behavior with human preferences \cite{dubey2024llama, gemma_2024,hejna2023few, narin2024evolutionary, gao2023scaling,zhong2025comprehensive}. Generally, RMs are trained on comparisons of outputs (chosen vs. rejected responses) to predict which output is better, often using data collected from human raters or AI assistants. Existing RMs can be categorized along two axes: (1) the model form \cite{lambert2024rewardbench} (discriminative RMs vs. generative RMs vs. implicit RMs) and (2) the feedback target (outcome-based vs. process-based supervision). Discriminative RMs \cite{cai2024internlm2technicalreport,yang2024regularizing,wang2024arithmetic} treat preference prediction as a binary (or scalar) regression problem: given an input and a candidate response, the model directly outputs a score (or probability of being the preferred response). By contrast, generative RMs use a language-model head to generate an evaluation or verdict based on a specific prompt \cite{zheng2023judging,lambert2024rewardbench,team2023gemini}, rather than directly outputting a numeric score. A third category, implicit RMs \cite{lambert2025tulu3pushingfrontiers}, effectively reparameterize preference learning within the model itself via Direct Preference Optimization~(DPO) \cite{rafailov2024direct}, enables construct preference pairs without requiring an explicit RM.
The second axis pertains to the nature of the feedback signal. Outcome-based RMs ~\cite{cobbe2021trainingverifierssolvemath} generate a single scalar reward for the entire response, reflecting its overall quality or correctness. Process-based RMs ~\cite{lightman2023letsverifystepstep} assess intermediate steps within a response, producing a sequence of rewards that reflect the quality of reasoning throughout the generation.

\noindent\textbf{Reward Models for Vision-Language Models.} Motivated by the observed benefits of preference-based alignment for VLMs \cite{wang2025enhancingreasoningabilitymultimodal}, the extension of reward modeling to the multimodal domain is an active research area. 
Most studies focused on generative RMs \cite{ouali2024clipdpovisionlanguagemodelssource,yu2024rlaifvopensourceaifeedback,xiong2025llavacriticlearningevaluatemultimodal} based on  open-source VLMs and explored data augmentation for response generation \cite{deng2024efficientselfimprovementmultimodallarge,deng2024enhancinglargevisionlanguage}. 
For discriminative RMs, LLaVA-RLHF \cite{sun2023aligninglargemultimodalmodels} pioneered the application of Reinforcement Learning from Human Feedback (RLHF) \cite{ouyang2022traininglanguagemodelsfollow} to VLMs by leveraging human feedback to train a multimodal RM. Building on this, they further proposed Fact-RLHF, which enhances reward signals by incorporating additional information (e.g., image captions).
To address the challenge of data scarcity, recent efforts have focused on the synthesis of large-scale, multi-domain preference datasets using advanced models as proxies for human evaluation. For instance, IXC-2.5-Reward \cite{zang2025internlm} leverages GPT-4o and verifier functions \cite{lambert2024rewardbench} to automatically generate preference labels, leading to significant performance gains. 

Most of the current approaches belong to ORM, providing reward signal for the final outputs of VLMs. More recently, Wang \textit{et al}. introduced VisualPRM \cite{wang2025visualprm}, a model trained on multimodal process supervision data that provides fine-grained multimodal process reward signals and serves as an effective critic model for test-time scaling of MLLMs, thereby enhancing the reasoning capabilities of existing VLMs.


\section{Method}\label{sec:method}

Our objective is to develop a multimodal reward model named Skywork-VL Reward, which takes as input an optional image, a textual prompt, and a candidate response generated by either a multimodal understanding model or a reasoning model. The model outputs a scalar reward score that reflects the quality or degree to which the response aligns with human preferences.
We achieve this by fine-tuning a pretrained VLM on a curated set of preference comparison data. In this section, we describe our dataset construction pipeline, the model architecture and modifications for reward modeling, the loss function used for pairwise preference training, and the overall training strategy.

\begin{figure*}
    \centering
    \includegraphics[width=1\linewidth]{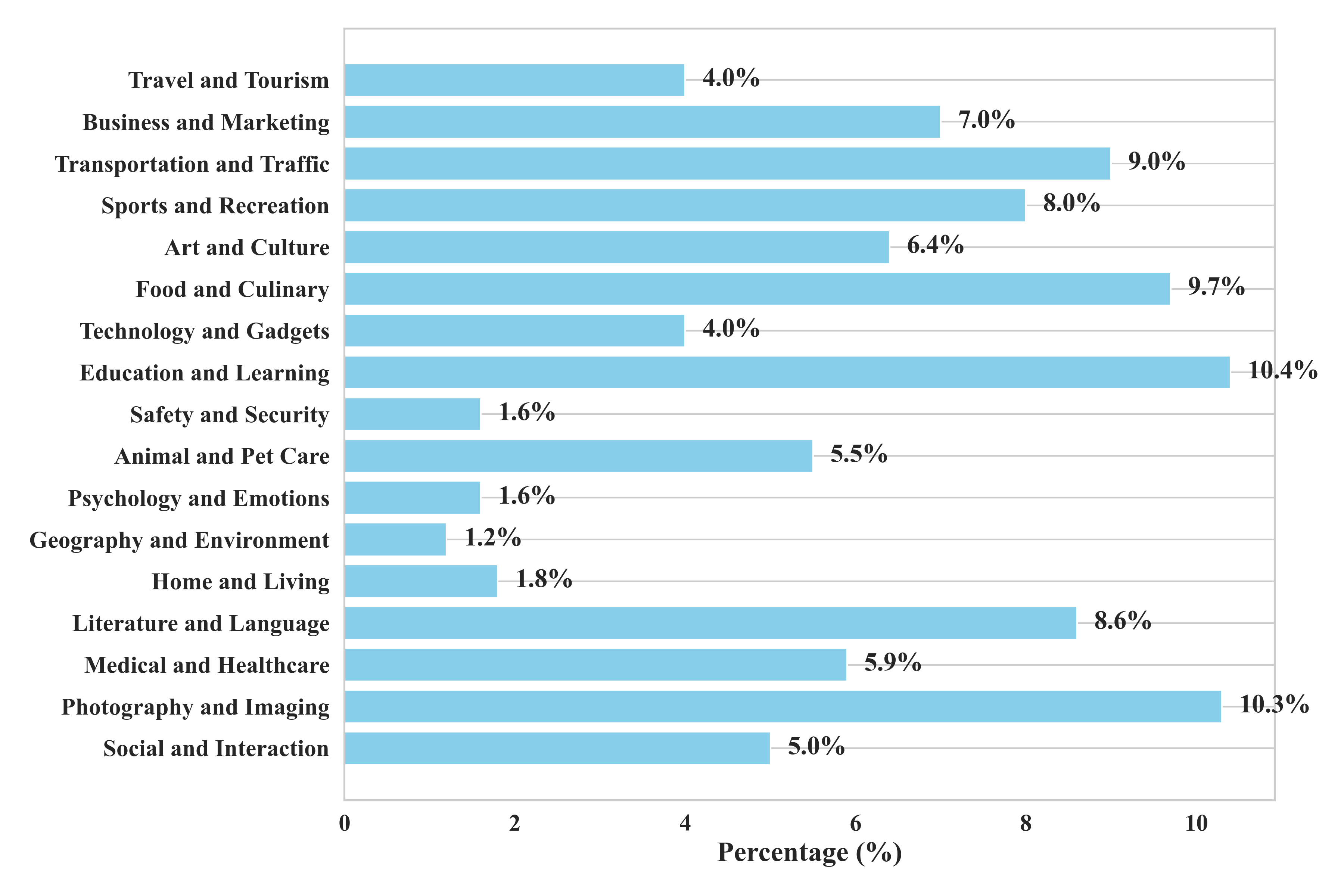}
    \caption{Distribution of Training Data from Open-Source Sources.} 
    \label{fig: data_1} 
\end{figure*} 

\subsection{Dataset Construction}\label{sec:method_data}

\paragraph{Open Source Data.} We construct a comprehensive training dataset for Skywork-VL Reward by integrating multiple open-source preference datasets and additional in-house annotations. The dataset primarily includes three sources: (1) LLaVA-Critic-113k \cite{xiong2025llavacriticlearningevaluatemultimodal}, (2) Skywork-Reward-Preference-80K-v0.2 \cite{liu2024skyworkrewardbagtricksreward}, and (3) RLAIF-V-Dataset \cite{yu2024rlaifvopensourceaifeedback}.
\begin{itemize}
    \item LLaVA-Critic-113k is an open-source dataset consisting of 113k multimodal instruction-response examples. Each example contains an image, a user query, and one or more model responses with associated quality judgments. This dataset uniquely provides both pointwise scores and pairwise rankings generated by GPT-4o, covering tasks from straightforward image descriptions to complex reasoning challenges. Each pair is often accompanied by explanatory annotations, enriching our understanding of judgment criteria.

\item Skywork-Reward-Preference-80K-v0.2 is a high-quality dataset comprising 80k pairs of human-preferred textual responses, covering diverse domains such as general Q\&A and creative writing. By eliminating noisy and inconsistent judgments through careful filtering, this dataset significantly enhances the text comprehension and alignment capabilities of Skywork-VL Reward, enabling it to effectively handle purely textual inputs.

\item RLAIF-V-Dataset is a large-scale multimodal feedback dataset containing 83,132 preference pairs. The instructions in this dataset are sourced from diverse datasets. Incorporating this dataset greatly enhances the general multimodal understanding abilities of Skywork-VL Reward, enabling robust performance across varied tasks and contexts.
\end{itemize}

\textbf{In-house Reasoning Data.} We augment existing datasets with a proprietary in-house dataset consisting of approximately 50,000 preference comparisons focused on complex reasoning tasks. The tasks primarily involved carefully curated multimodal problems spanning mathematics, physics, biology, and chemistry (Table \ref{tab:t1}). These comparisons were collected through human annotation, where annotators assessed the correctness and reasoning quality of various VLM-generated reasoning-style responses. 

\subsection{Data Curation Procedure}
Our data curation involved three stages. Open-source datasets were utilized in the initial two stages. For the final stage, we employed a reasoning dataset derived from internal human annotation resources.

\paragraph{Stage 1:} The first data curation stage involved deduplication and filtering of the aggregated dataset. Specifically, our data curation involved the following:

\begin{itemize}
    \item \textbf{Deduplication:} Removal of identical pairs across different sources.
    \item \textbf{Similarity Filtering:} Elimination of highly similar samples based on semantic similarity.
    \item \textbf{Judgment Filtering:} Discarding pairs with ambiguous or low-confidence preference judgments (compared to GPT-4o), where "equal quality" was considered ambiguous.
\end{itemize}
    
This process yielded approximately 200,000 distinct, high-confidence preference pairs. This refined dataset was subsequently used to train a surrogate RM, which then scored the entire dataset.

\paragraph{Stage 2:} 
We further refined the collected dataset based on scores predicted by the surrogate RM in the second stage, with the following principles:

\begin{itemize}

\item \textbf{Low-Scoring Chosen Responses:} If the chosen response in a preference pair received a low score from the surrogate RM, we regenerated that response using GPT-4o, effectively replacing the original.
\item \textbf{Minimal Score Difference:} For preference pairs where the score difference between the chosen and rejected responses was minimal, we also regenerated the chosen response with GPT-4o to enhance clarity.

\end{itemize}

Following this refinement process, a total of 150,000 data samples were retained. The distribution of this resulting training data is presented in Figure \ref{fig: data_1}.

\begin{table}[t]
\centering
\small
\caption{Distribution of In-house Training Data.}
\begin{tabular}{lccccc}
\toprule
Field & Mathematics & Physics & Biology & Chemistry & Others \\
\midrule
Percentage (\%) & 35.4 & 24.6 & 14.7 & 20.2 & 5.1 \\
\bottomrule
\end{tabular}
\label{tab:t1}
\end{table}

\begin{table}[t]
\centering
\small
\caption{Percentage of Generation Approaches of In-house Training Data.}
\begin{tabular}{lcc}
\toprule
Approach & Direct generation & Two-step generation \\
\midrule
Percentage (\%) & 47.4 & 52.6 \\
\bottomrule
\end{tabular}
\label{tab:t2}
\end{table}

\paragraph{Stage 3:} In this stage, reasoning-style responses were generated via two distinct methods, with the percentage breakdown presented in Table \ref{tab:t2}:

\begin{itemize} 
\item \textbf{Direct generation:} Using our vision-language reasoning model Skywork R1V to directly produce reasoning-style responses conditioned on the original multimodal input.

\item \textbf{Two-step generation:} First employ a standard VLM model (we use the InternVL series \cite{chen2024expanding}) to generate a textual description of the image, which serves as a substitute for the original visual input. Then, this textual description is fed into llm-reasoner (we use the Deepseek R1 series \cite{deepseekai2025deepseekr1incentivizingreasoningcapability}) to generate the final reasoning-style response. \end{itemize}

After the above procedure, our final training dataset consists of approximately 190k comparison samples (each a prompt with two responses and a preference), of which about 70\% include images. We believe this dataset is one of the largest and most diverse multimodal preference collections to date, and it is a key enabler of Skywork-VL Reward's strong generalization.

\subsection{Model Architecture}\label{sec:method_model}

Our RM is built upon the Qwen2.5-VL-7B-Instruct \cite{bai2025qwen2}. Qwen2.5-VL-7B-Instruct is a 7-billion-parameter VLM recently released by Alibaba Cloud, which extends the Qwen-7B LLM with visual understanding capabilities. The architecture consists of three main components: a Vision Transformer (ViT) as the visual encoder, a vision-language adapter/projector, and the Qwen language model decoder.
The visual encoder processes an input image into a sequence of patch features. These raw image features are then passed through a small adapter module that projects them into a compact set of visual tokens that the language model can attend to. The language model  then takes these visual tokens, along with text tokens from the prompt, and generates a response.

For Skywork-VL Reward, we modify this architecture to output a scalar reward instead of a text sequence. We remove Qwen's causal LM head used for token prediction and replace it with a reward head that produces a single score. Concretely, we attach a fully-connected layer on top of the final hidden state to predict the reward.The reward head processes the final hidden state following the answer's last token, producing a raw score $r_{\theta}$. This score is used during training to calculate a preference loss across various answers. At inference, the Skywork-VL Reward model evaluates a given prompt and response by outputting a quality score.

\subsection{Reward Model Loss Function}\label{sec:method_loss}

We train Skywork-VL Reward using a standard pairwise preference loss \cite{bradley1952rank} commonly used in the reward modeling stage of RLHF. For a given comparison example, we have a prompt (with optional image) $x$, a preferred response $y^{+}$ (the one chosen by the annotator or judge), and a dispreferred response $y^{-}$ (the one that was rejected). The model generates a scalar reward score for each response: $s^{+} = r_\theta(x, y^{+})$ for the preferred answer and $s^{-} = r_\theta(x, y^{-})$ for the dispreferred one, where $r_\theta$ represents the reward model's output. Our loss function aims to maximize the difference between $s^{+}$ and $s^{-}$, which can be formulated as:

\begin{equation}\label{eq:loss}
\mathcal{L}_{\text{RM}}(\theta) \;=\; -\log \sigma\!\Big( r_\theta(x, y^{+}) \,- r_\theta(x, y^{-}) \Big)\,,
\end{equation}
where $\sigma(z) = \frac{1}{1+e^{-z}}$ is the sigmoid function.  This formulation focuses solely on learning the relative ranking of responses, encouraging the model to assign higher scores to preferred answers without explicitly calibrating the absolute reward scores.
Consequently, examples with equal or near-equal preferences are prone to introducing ambiguity and were therefore excluded from our training data.

\section{Experiment} 

\subsection{Training Details}\label{sec:method_training} 
\textbf{Training Parameter.} To efficiently fine-tune the model as a reward scorer, we adopt a partial parameter freezing strategy \cite{yang2024regularizinghiddenstatesenables}. In particular, we freeze the entire visual encoder of Qwen2.5-VL-7B-Instruct  to preserve its visual abilities pretrained on massive image-text data. The trainable weights in Skywork-VL Reward are limited to the projector, language backbone, and the reward head.

\textbf{Two-Stage Fine-Tuning Procedure.} 
We formulate preference learning as a supervised learning task over the constructed preference dataset. The fine-tuning follows a two-stage training strategy. In the first stage, the model is trained exclusively on multimodal preference data, allowing it to develop strong vision-language alignment capabilities. In the second stage, we additionally incorporate pure-text preference data to further improve the model's generalization and reasoning abilities in text-only scenarios.

We use AdamW \cite{loshchilov2019decoupledweightdecayregularization} with a moderate learning rate for optimization in the first stage ($10^{-5}$) and a lower learning rate in the second stage ($10^{-6}$ ). And the model is fine-tuned for 2 epochs per stage, which we find sufficient for convergence.

\subsection{Evaluation Benchmarks}

We evaluate Skywork-VL Reward on two benchmarks: VL-RewardBench \cite{li2024vlrewardbenchchallengingbenchmarkvisionlanguage} and RewardBench \cite{lambert2024rewardbench}.
VL-RewardBench is designed to assess vision-language reward modeling. It contains 1,250 carefully curated examples spanning general multimodal queries, visual hallucination detection, and complex reasoning tasks involving images. 
RewardBench is a pure-text benchmark targeting reward functions for language models. This dataset includes prompt-chosen-rejected triplets covering a diverse range of topics within general chat, safety, and reasoning. We report the performance on both benchmarks, specifically focusing on their key evaluation dimensions and the aggregated overall accuracy.
\subsection{Baselines}

For VL-RewardBench, we compare Skywork-VL Reward against a broad range of RMs, including both cutting-edge proprietary models and leading open-source alternatives.
The proprietary multimodal RMs (closed-source) in our evaluation include GPT-4o \cite{openai2023gpt4}, Claude 3.5 \cite{Claude2024} with Vision, and Google Gemini 1.5 \cite{team2023gemini}. These models represent top-performing industrial models and serve as upper-bound references for RM performance.
The involved prominent open-source models include Qwen2-VL-7B-Instruct \cite{Qwen2VL}, MAmmoTH-VL-8B \cite{guo2024mammothvlelicitingmultimodalreasoning}, Qwen2.5-VL-7B-Instruct \cite{bai2025qwen2}, InternVL3-8B \cite{zhu2025internvl3exploringadvancedtraining}, Qwen2-VL-72B-Instruct \cite{Qwen2VL}, IXC-2.5-Reward-7B \cite{zang2025internlm}, Molmo-72B \cite{deitke2024molmopixmoopenweights}, QVQ-72B-Preview \cite{qvq-72b-preview}, Qwen2.5-VL-72B-Instruct \cite{bai2025qwen2}, and InternVL3-78B \cite{zhu2025internvl3exploringadvancedtraining}. 

For RewardBench, we evaluate several advanced language-only RMs, including InternLM2-7B-Reward \cite{cai2024internlm2technicalreport}, Skywork-Reward-Llama3.1-8B \cite{liu2024skyworkrewardbagtricksreward}, Skywork-Reward-Llama3.1-8B-v0.2 \cite{liu2024skyworkrewardbagtricksreward}, and QRM-Llama3.1-8B-v2 \cite{dorka2024quantile}. We also evaluate multimodal RMs that are comparable in size to our own models including Qwen2-VL-7B-Instruct, InternVL3-8B, IXC-2.5-Reward-7B, and Qwen2.5-VL-7B-Instruct.

In our experiments, Qwen2.5-VL-7B-Instruct, InternVL3-8B, Qwen2.5-VL-72B-Instruct, IXC-2.5-Reward-7B, and InternVL3-78B were reproduced by ourselves, whereas the results for the remaining models were obtained from official reports.

\begin{table*}[t]
    \centering
    \small 
    \caption{Evaluation Results on VL-RewardBench.}
    \resizebox{.95\textwidth}{!}{
    \begin{tabular}{@{}lcccccc@{}}
    \toprule 
    \textbf{Models}  & \textbf{Model Size}  &  \textbf{General}  & \textbf{Hallucination} & \textbf{Reasoning} & \textbf{Overall Accuracy}  & \textbf{Macro Average} \\
    \midrule 
    \multicolumn{7}{c}{\emph{Proprietary Models}} \\ 
    \midrule 
    Claude-3.5-Sonnet(2024-06-22) &-&43.4&55.0&62.3&55.3&53.6\\ 
    Gemini-1.5-Flash (2024-09-24) &-&47.8&59.6&58.4&57.6&55.3\\
    GPT-4o(2024-08-06) &-&49.1&67.6&70.5&65.8&62.4\\
    Gemini-1.5-Pro(2024-09-24) &-&50.8&72.5&64.2&67.2&62.5\\
    Gemini-2.0-flash-exp(2024-12) &-&50.8&72.6&70.1&{68.8}&{64.5}\\
    \midrule
    \multicolumn{7}{c}{\emph{Open-Source Models}} \\ 
    \midrule
    Qwen2-VL-7B-Instruct  &7B&31.6&19.1&51.1&28.3&33.9 \\
    MAmmoTH-VL-8B &8B&36.0&40.0&52.0&42.2&42.7 \\
    Qwen2.5-VL-7B-Instruct &7B&43.4&42.0&63.0&48.0&49.5 \\
    InternVL3-8B&8B&60.6&44.0&62.3&57.0&55.6 \\
    IXC-2.5-Reward-7B&7B&80.3&65.3&60.4&66.3&68.6 \\
    Qwen2-VL-72B-Instruct &72B&38.1&32.8&58.0&39.5&43.0 \\
    Molmo-72B-0924 &72B&33.9&42.3&54.9&44.1&43.7 \\
    QVQ-72B-Preview &72B&41.8&46.2&51.2&46.4&46.4 \\
    Qwen2.5-VL-72B-Instruct &72B&47.8&46.8&63.5&51.6&52.7 \\
    InternVL3-78B &78B&67.8&52.5&64.5&63.3&61.6 \\
   \midrule
  
   Skywork-VL Reward (Ours)& 7B & {66.0} & {80.0} & 61.0 & \textbf{73.1} & \textbf{69.0} \\
    \bottomrule
    \end{tabular}}
    \label{tab:vl_reward}
\end{table*}

\begin{table*}[t]
    \centering
    \small 
    \caption{Evaluation Results on RewardBench.}
    \resizebox{.7\textwidth}{!}{
    \begin{tabular}{@{}lccccc@{}}
    \toprule
    \textbf{Models} & \textbf{Chat} & \textbf{Chat Hard} &  \textbf{Safety} & \textbf{Reasoning} & \textbf{Avg Score} \\
    \midrule
    \multicolumn{6}{c}{\emph{Language-Only Reward Models}} \\
    \midrule
    InternLM2-7B-Reward & {99.2} & 69.5 & 87.2 & 94.5 & 87.6 \\
    Skywork-Reward-Llama3.1-8B & 95.8 & 87.3 & 90.8	 & 96.2	 & 92.5 \\
    Skywork-Reward-Llama-3.1-8B-v0.2 & 94.7 & {88.4} & {92.7} & 96.7 & 93.1 \\
    QRM-Llama3.1-8B-v2 & 96.4 & 86.8 & 92.6 & 96.8 & \textbf{93.1} \\
    \midrule
    \multicolumn{6}{c}{\emph{MultiModal Reward Models}} \\
    \midrule
    Qwen2-VL-7B-Instruct & 65.1 & 50.9 & 55.8 & 68.3 & 60.0 \\
    InternVL3-8B & {97.2} & 50.4 & 83.6 & 83.9 & 78.8 \\
    Qwen2.5-VL-7B-Instruct & 94.3 & 63.8 & 84.1 & 86.2 & 82.1 \\
    IXC-2.5-Reward-7B  & 90.8 & 83.8 & 87.8 & 90.0 & 88.1 \\
    \midrule
    Skywork-VL Reward (Ours)& 90.0 & {87.5} & {91.1} & 91.8 & \textbf{90.1} \\
    \bottomrule
    \end{tabular}}
    \label{tab:reward-bench}
\end{table*}

\addtocounter{footnote}{1}
\subsection{VL-RewardBench Evaluation}

Table~\ref{tab:vl_reward} presents comparisons with both proprietary and open-source models on VL-RewardBench.

In the general category, skywork-VL Reward achieves a score of 66.0\%, significantly outperforming even the strongest proprietary model, Gemini-2.0-flash-exp (50.8\%). However, a gap remains compared to IXC-2.5-Reward-7B (80.3\%).
In the hallucination category, our model achieves the best score (80.0\%), surpassing both proprietary models (e.g., Gemini-2.0-flash-exp at 72.6\%) and the top-performing open-source model, IXC-2.5-Reward-7B (65.3\%). This result highlights our model’s strong capability in mitigating factual inconsistencies.
Our model also demonstrates robust performance in the reasoning category. It achieves a reasoning score of 61.0\%, which is comparable to that of the much larger InternVL3-78B (64.5\%), despite having 10× fewer parameters.

Our model achieves an overall accuracy of 73.1\% and a Macro Average of 69.0\%, demonstrating superior performance across diverse task types and surpassing the best proprietary model, Gemini-2.0-flash-exp (68.8\% overall accuracy and 64.5\% Macro Average), and the second-best open-source model, IXC-2.5-Reward-7B (66.3\% overall accuracy and 68.6\% Macro Average). These results demonstrate the effectiveness of our method in providing reliable reward signals for multimodal tasks.

\begin{figure*}
    \centering
    \includegraphics[width=1\linewidth]{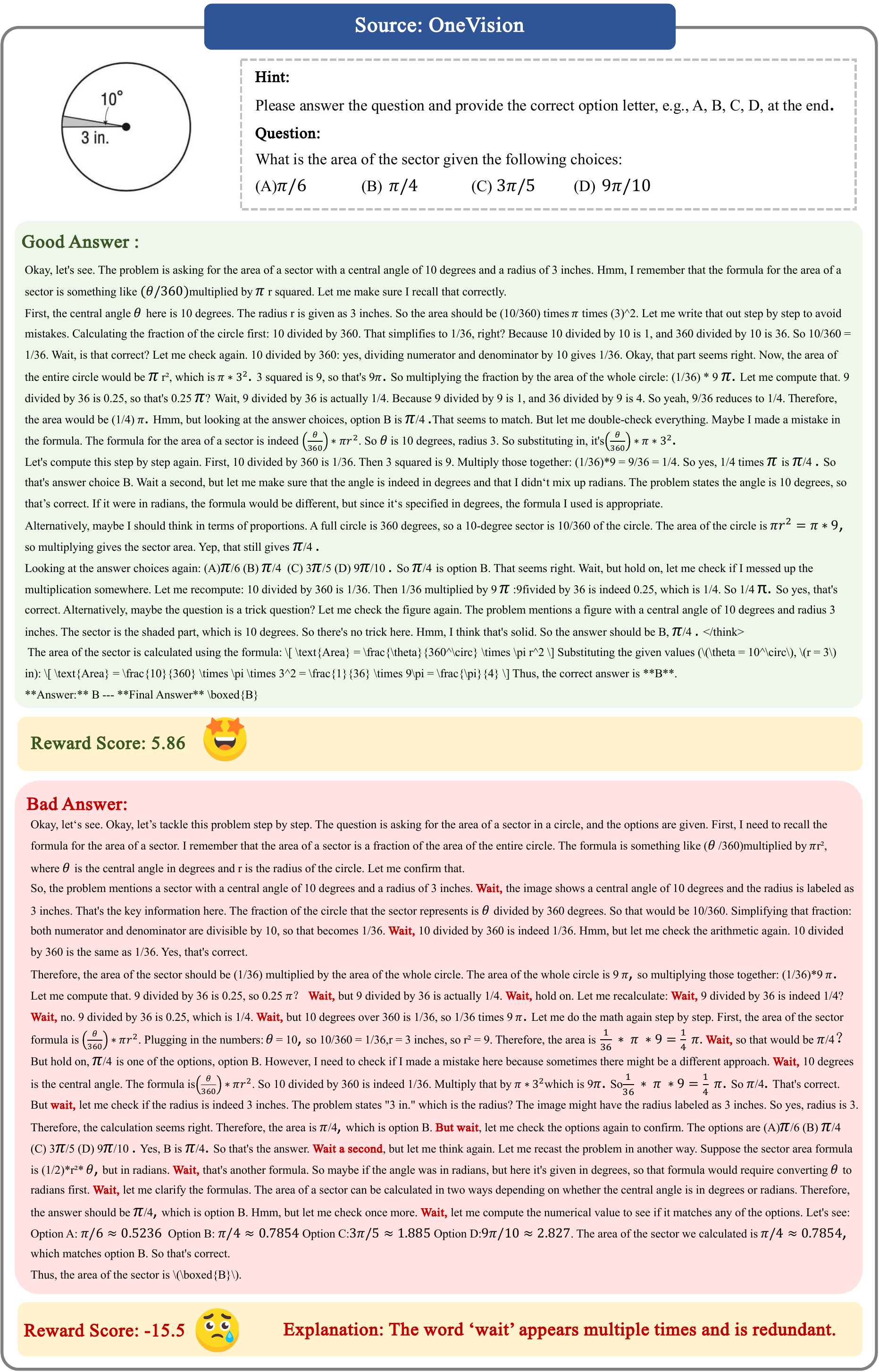}
    \caption{Evaluating Skywork R1V on Mathematical Problems.} 
    \label{fig:case3} 
\end{figure*} 

\begin{figure*}
    \centering
    \includegraphics[width=1\linewidth]{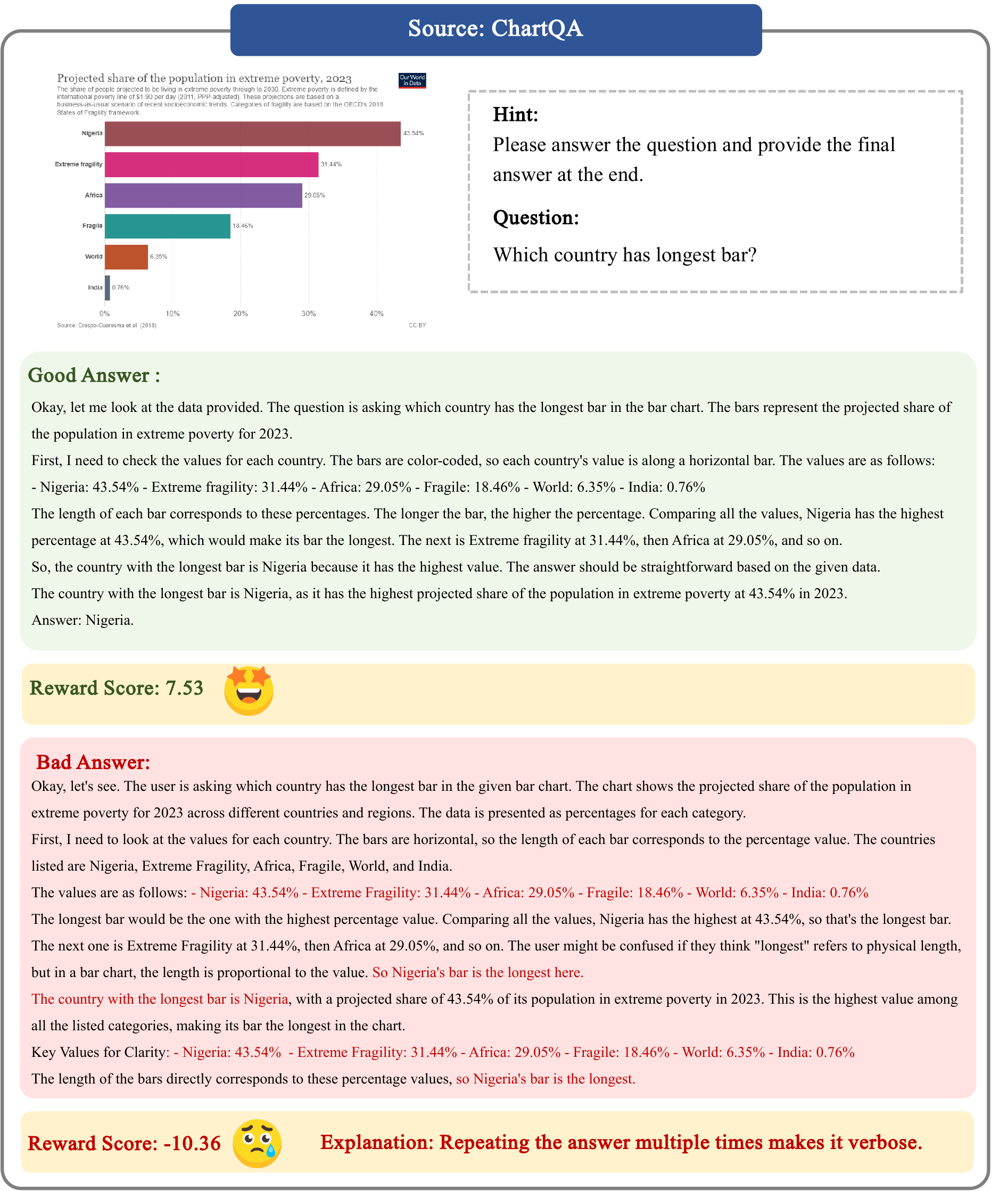}
    \caption{Evaluating Skywork R1V on Chart Problems.} 
    \label{fig:case2} 
\end{figure*}

\subsection{RewardBench Evaluation}

Table~\ref{tab:reward-bench} reports the results on RewardBench, a language-focused reward benchmark.

Our model achieves an average score of 90.1\% on RewardBench, achieve advanced performance among all open-source multimodal RMs of comparable scale and outperforming the second-best model, IXC-2.5-Reward-7B, by 2.0\%. It also shows competitive performance against advanced language-specific RMs, such as QRM-Llama3.1-8B-v2 (93.1\%).

Compared to other multimodal RMs of similar size, our model leads in Chat Hard (87.5\%), Safety (91.1\%), and Reasoning (91.8\%), outperforming the second-best models by 3.7\%, 3.3\%, and 1.8\%, respectively. These results highlight the robustness and well-rounded performance of our model across challenging and safety tasks. Furthermore, the results validate that our model is not only effective in handling multimodal data but also exhibits strong capabilities on pure-text inputs.

\subsection{Case Study}
\label{sec:case_studies}

We present two illustrative examples that highlight the efficacy of our Skywork-VL Reward across distinct reasoning scenarios.  
For each example, we supply a multimodal prompt together with a \emph{good} and a \emph{bad} answer. For each given answer, Skywork-VL Reward produces a scalar reward value upon querying.

The first example (Figure \ref{fig:case3}) is a geometry problem that asks for the area of a circular sector.  
While both candidate answers arrive at the same numerical conclusion, the good answer showcases accurate reasoning in its derivation, unlike the bad answer's rambling, self-corrective approach. This indicates that Skywork-VL Reward strongly favors the concise solution, demonstrating its sensitivity to the quality of reasoning rather than just the final correctness.

The second example (Figure \ref{fig:case2}), involving identifying the country with the longest bar in an extreme poverty rate chart, further illustrates this. The good answer concisely states the label and cites relevant percentages, while the bad answer redundantly lists the same numbers. Again, Skywork-VL Reward strongly favors the compact explanation, demonstrating robustness across a different visual domain. These two cases, spanning distinct domains, highlight Skywork-VL Reward's consistent ability to differentiate well-structured reasoning from verbose or confused discourse. The significant reward gap observed in both settings suggests the model captures a valuable alignment signal for downstream reinforcement learning.

\begin{table}[t]
\centering
\small 
\caption{Performance Evaluation on the MathVista Benchmark Using MPO with Different Reward Models. }
\begin{tabular}{lcccc}
\toprule
Model & Base & Qwen2.5-VL-7B-Instruct  & InternVL3-8B & Ours \\
\midrule
Performance (\%) & 69.2 & 71.2 & 71.8 &  73.5 \\
\bottomrule
\end{tabular}
\label{tab:mpo}
\end{table}
\subsection{Skywork-VL Reward for Mixed Preference Optimization}
We examine the effect of Skywork-VL Reward as a reward signal for MPO \cite{wang2025enhancingreasoningabilitymultimodal}, a recent strategy to further improve model alignment. MPO refers to optimizing the behavior of a model using a mixture of preference signals rather than a single RM.  This approach was proposed to stabilize and enhance training, especially for complex reasoning tasks that benefit from diverse feedback. 

Moreover, preference data generated using our Skywork-VL reward demonstrates high effectiveness in training MPO, resulting in substantial gains in multimodal reasoning abilities.
We leverage Skywork-VL Reward to generate preference data, which is subsequently used to fine-tune a VLM reasoner (the base model of Skywork R1V2 \cite{chris2025skyworkr1v2multimodalhybrid}) via MPO. Performance is evaluated on MathVista \cite{lu2023mathvista}, a challenging benchmark for mathematical reasoning over visual content. As shown in Table~\ref{tab:mpo}, incorporating Skywork-VL Reward as an additional reward yields a notable improvement: the model’s MathVista score increases from 69.2\% to 73.5\%. This improvement demonstrates the potential of Skywork-VL Reward as a critical component in training VLMs capable of long-CoT reasoning.

\section{Conclusion}\label{sec:conclusion}
In this work, we introduce Skywork-VL Reward, a multimodal reward model for VLMs, aimed at addressing the critical need for reliable and general-purpose evaluators in multimodal understanding and reasoning tasks. Through the construction of a large-scale, meticulously curated preference dataset encompassing various tasks and scenarios, coupled with a two-stage training paradigm,  our model is able to effectively assess responses generated by both standard VLMs and VLM reasoners.  Empirical results demonstrate the state-of-the-art performance of Skywork-VL Reward on the VL-RewardBench benchmark and its competitive capabilities on the text-only RewardBench. Furthermore, integrating Skywork-VL Reward to provide supervised signals for MPO significantly enhances the multimodal reasoning abilities of VLMs, highlighting its practical value.

\bibliography{main}
\bibliographystyle{unsrt}
\end{document}